\documentclass[10pt, a4paper]{article}

\usepackage[final]{lrec2026} 

\usepackage{amsmath}
\usepackage{multirow}
\usepackage{comment}
\usepackage{booktabs}

\title{Social Meaning in Large Language Models: Structure, Magnitude, and Pragmatic Prompting}

\name{Roland M\"uhlenbernd} 

\address{Leibniz-Centre General Linguistics, Berlin, Germany \\
         muehlenbernd@leibniz-zas.de\\}

\abstract{
Large language models (LLMs) increasingly exhibit human-like patterns of pragmatic and social reasoning. This paper addresses two related questions: do LLMs approximate human social meaning not only qualitatively but also quantitatively, and can prompting strategies informed by pragmatic theory improve this approximation? To address the first, we introduce two calibration-focused metrics 
distinguishing structural fidelity from magnitude calibration: 
the Effect Size Ratio (ESR) and the Calibration Deviation Score (CDS). To address the second, we derive prompting conditions from two pragmatic assumptions: that social meaning arises from reasoning over linguistic alternatives, and that listeners infer speaker knowledge states and communicative motives. Applied to a case study on numerical (im)precision across three frontier LLMs, we find that all models reliably reproduce the qualitative structure of human social inferences but differ substantially in magnitude calibration. Prompting models to reason about speaker knowledge and motives most consistently reduces magnitude deviation, while prompting for alternative-awareness tends to amplify exaggeration. Combining both components is the only intervention that improves all calibration-sensitive metrics across all models, though fine-grained magnitude calibration remains only partially resolved. LLMs thus capture inferential structure while variably distorting inferential strength, and pragmatic theory provides a useful but incomplete handle for improving that approximation.\\ \newline \Keywords{large language models, social inference, pragmatics, social meaning,
magnitude calibration, pragmatic prompting, evaluation, numerical (im)precision}}

\begin{document}

\maketitleabstract

\section{Introduction}

Large language models (LLMs) increasingly exhibit sophisticated forms of
pragmatic and social reasoning. Recent work has shown that they can recover
conversational implicatures \citep{Ruis2024, Sravanthi2024,Scherrer2024}, reason
pragmatically about scalar expressions \citep{Cho2024}, and produce
context-sensitive social judgments that align with expert human evaluations
\citep{Kieser2024}. A growing body of work further suggests that LLMs can
simulate human samples in social science experiments, reproducing population-level
patterns of social judgment \citep{Argyle2023, Santurkar2023}. This paper
pursues two related but distinct questions about the quality of this reasoning.

The first concerns measurement. Most existing evaluations of LLM social
reasoning focus on directional or categorical agreement: whether a model
identifies the correct implication or ranks alternatives similarly to humans.
Yet many aspects of human social evaluation are inherently graded. The strength
of inferred traits (e.g., competence, friendliness) depends on subtle
interactions between linguistic form and context. A model may reproduce the
qualitative direction of an effect while systematically exaggerating or
attenuating its magnitude. This structure--magnitude dissociation has been
documented in broad social science domains, where LLMs have been shown to
reproduce effect directions while overestimating magnitudes by factors of 2--10
\citep{Hewitt2024, Cui2025, Argyle2023}. Crucially, however, most
existing work reports this discrepancy as a descriptive side finding, without
metrics designed to quantify it in a principled way \citep{Hullman2025}. We
address this gap by introducing two magnitude-sensitive metrics, the
\emph{Effect Size Ratio (ESR)} and the \emph{Calibration Deviation Score
(CDS)}, that operationalize the distinction between \emph{structural
fidelity} and \emph{magnitude calibration}.

The second question concerns explanation and intervention. Can prompting
strategies informed by pragmatic theory improve how well LLMs approximate
human social meaning? We ground our prompting conditions in two well-established
assumptions from pragmatics: that social meaning arises from reasoning over
linguistic alternatives in context, and that listeners evaluate speakers by
inferring their knowledge states and communicative motives. If LLMs engage in
genuinely pragmatic social reasoning, then prompts that explicitly activate
these reasoning processes should modulate model behavior in theoretically
predictable ways, which allows us to test not only how well LLMs approximate
human social meaning, but whether pragmatic theory provides a useful handle for
improving that approximation.

We apply both contributions to a case study on numerical (im)precision, a
domain in which the interplay between linguistic form, context, and social
inference is well documented \citep{BeltramaSoltBurnett2022,
SoltEtAl2025}, and in which LLM pragmatic behavior has already
attracted attention \citep{Tsvilodub2025}. Across three frontier LLMs and four
theory-motivated prompting conditions, we find a consistent dissociation: all
models achieve high structural alignment but differ markedly in magnitude
calibration. Knowledge-and-Motives-Aware prompting partially restores human-like
calibration in overconfident models, while combined prompting yields
architecture-dependent trade-offs rather than uniform improvement.

\section{Theoretical Background}
\label{sec:background}



Many instances of social meaning are not directly encoded in linguistic form
but emerge from inferential processes listeners apply when interpreting
speakers' utterances \citep{Acton2019,Beltrama2020}. These inferences
often concern social attributes of the speaker, including competence,
knowledgeability, and communicative intent, that listeners
update based on the speaker's linguistic choices and the context in which
they occur \citep{BeltramaPapafragou2023, BeltramaSoltBurnett2022,
SoltEtAl2025}. Two well-established assumptions from pragmatics ground
our evaluation framework and motivate our prompting conditions.

\paragraph{Reasoning over Alternatives.}
Listeners evaluate a speaker's linguistic choice against alternatives they
could have produced. In Gricean pragmatics \citep{Grice1975}, a speaker's
selection of a weaker or less precise expression where a stronger one was
available licenses inferences about their epistemic state or intent
\citep{Levinson2000}. Alternative-sensitive reasoning produces context-dependent
social evaluations: the social meaning of numerical precision is modulated by
contextual demands, with precise forms enhancing perceived status more strongly
in high-precision contexts (e.g., formal testimony) than in casual ones
\citep{BeltramaSoltBurnett2022,SoltEtAl2025}. Social inference is thus not
triggered by form alone, but by the relationship between form, available
alternatives, and context.

\paragraph{Speaker Knowledge and Motives.}
Listeners also infer social attributes by reasoning about \emph{why} a speaker
chose a particular expression: what knowledge states and communicative
motives plausibly explain the observed choice. This is central to Grice's
\citeyearpar{Grice1957} account of meaning as intention recognition, and is
grounded in the notion that utterances are interpreted against a shared
communicative context \citep{Stalnaker1999}. Empirically,
\citet{BeltramaPapafragou2023} showed that violations of Gricean norms of
relevance and informativeness systematically reduce social evaluations of
competence and warmth, mediated by listeners' inferences about speaker motives.

\paragraph{RSA as a Unifying Framework.}
The Rational Speech Act framework \citep{FrankGoodman2012, GoodmanFrank2016}
is the most prominent formal account within the broader tradition of
probabilistic pragmatics \citep{FrankeJaeger2016}, and integrates both
assumptions above. In RSA, a pragmatic listener interprets an utterance by
reasoning jointly over the space of alternative utterances a rational speaker
could have produced \emph{and} over the speaker's latent beliefs and
communicative goals. Social meaning emerges from this joint inference: the
same form (e.g., an approximate number) can warrant different social
evaluations depending on which alternatives were available and what knowledge
state or motive best explains the speaker's choice. This integration motivates
treating the two assumptions not as independent factors but as complementary
components of a single inferential process, a structure directly reflected
in our Combined prompting condition.

\paragraph{Implications for Evaluation and Prompting.}
These two assumptions have direct methodological consequences. First, they
imply that evaluating social meaning in LLMs requires going beyond directional
agreement: a model may reproduce the \emph{direction} of a social inference
while failing to capture its graded \emph{strength}, which depends on how
competing alternatives and inferred speaker states are weighted. This motivates
our distinction between structural fidelity and magnitude calibration, and the
metrics we introduce to operationalize it.

Second, the assumptions motivate our prompting conditions directly. If social
inference involves reasoning over alternatives and epistemic uncertainty about
potential knowledge states and motives of the speaker, then prompts that
explicitly activate these two aspects should modulate model behavior in
theoretically predictable ways --- allowing us to test not only how well LLMs
approximate human social meaning, but whether pragmatic theory provides a
useful handle for improving that approximation.

\section{Behavioral Baseline: Social Inferences from (Im)Precision}
\label{sec:human}


We ground our LLM evaluation in Experiment~1 of \citet{SoltEtAl2025}, which
investigates how the choice of numerical precision level conveys social meaning
about the speaker, and how this meaning is modulated by the pragmatic
requirements of the utterance context. The study's central question is whether
the degree to which the level of precision in an expression impacts attributions
of competence, knowledgeability, and related traits depends on the contextual
demands for precision, or whether it operates uniformly regardless of context.
This tests the core pragmatic prediction that social meaning is not a fixed
property of linguistic form but arises from the relationship between form and
context. Numerical (im)precision is a particularly well-suited test case for
LLM evaluation: it offers a clearly defined set of linguistic alternatives
(precise vs.\ approximate forms), experimentally validated human effect sizes
as a quantitative benchmark, and prior evidence that LLMs engage in
precision-related pragmatic reasoning \citep{Tsvilodub2025}.

\paragraph{Design and materials.}
Participants ($N = 371$) were recruited online and randomly assigned to one of
six everyday dialog scenarios involving a numerical expression, in one of four conditions, crossing utterance form (precise vs.\ approximate numerical expression) with contextual precision requirements (high-precision [HP] vs.\ low-precision [LP] needs).
Scenarios were pretested to ensure that their two contextual versions differed
reliably in required precision level. For each scenario, participants rated the
speaker on six social dimensions using 7-point Likert scales:
\textit{competent}, \textit{knowledgeable}, \textit{well-prepared}
(competence-related); \textit{helpful}, \textit{likeable}
(likeability-related); and \textit{pedantic}.
Table~\ref{tab:example} illustrates the 2$\times$2 design using the
\textit{bicycle} scenario. The HP and LP contexts establish different
pragmatic demands for precision, such that the social cost of using an
approximate form is predicted to be higher when precision is situationally
required.

\begin{table}[t!]
\centering
\small
\begin{tabular}{p{2.0cm} p{2.0cm} p{2.3cm}}
\toprule
& \textbf{Precise} & \textbf{Approximate} \\
\midrule
\textbf{HP}  (insurance claim) &
\textit{``The bicycle cost \$500.''} &
\textit{``The bicycle cost about \$500.''} \\
\addlinespace
\textbf{LP}  (friend inquiry) &
\textit{``The bicycle cost \$500.''} &
\textit{``The bicycle cost about \$500.''} \\
\bottomrule
\end{tabular}
\caption{Example stimuli from the \textit{bicycle} scenario across the
four experimental conditions. HP context: Jamie reports the cost to an
insurance agent. LP context: Jamie answers a friend who is casually 
considering buying a bicycle. The utterance form
(precise vs.\ approximate) is identical across contexts; only the
pragmatic demands differ.}
\label{tab:example}
\end{table}

\paragraph{Results.}
The study yielded ten statistically significant effects that constitute the
directional structure of the human data. Five are \emph{main effects of
form}: precise speakers were rated significantly higher than approximate
speakers on \textit{competent}, \textit{knowledgeable}, \textit{well-prepared},
 \textit{helpful} and pedantic (all $p < .001$). 
Five additional effects are \emph{form $\times$ context interactions}:
the rating advantage of precise over approximate was significantly larger
in HP than LP for \textit{competent}, \textit{well-prepared},  \textit{helpful}, \textit{likeable} ($p < .001$) and \textit{knowledgeable} ($p < .05$). Together,
these effects reflect the context-sensitivity of social meaning: the social
cost of imprecision is more pronounced when precision is situationally
required, while the social benefit of approximation emerges most clearly when
high precision is not called for.

These ten effects (five main effects and five interactions) define the
benchmark against which LLM outputs are evaluated in Section~\ref{sec:results}.

\section{LLM Evaluation}
\label{sec:llm}

\paragraph{Models and protocol.}
We evaluated three frontier LLMs accessed via API:
\begin{itemize}
    \item GPT (\texttt{gpt-4o-mini})
    \item Claude (\texttt{claude-sonnet-4-20250514})
    \item Gemini (\texttt{gemini-2.5-pro})
\end{itemize}
For each combination of scenario, context, utterance form, and social
attribute, models were prompted to rate the speaker on the given attribute
using the identical 7-point scale as in the human experiment. Each query
was run $n = 10$ times at temperature $\tau = 1.0$, and model outputs were
averaged to compute mean ratings per attribute $\times$ context $\times$
form condition, matching the structure of the human dataset.

\paragraph{Prompting conditions.}
To probe the role of pragmatic reasoning in LLM social inference, we
implemented four prompting regimes grounded in the theoretical distinctions
introduced in Section~\ref{sec:background}.
\begin{itemize}
\item \emph{Minimal (MIN):} Reflects the exact instructions of the human
experiment, serving as the baseline for default inference behavior. An
example of a full prompt is provided in Appendix~\ref{appxA}.
\item \emph{Alternative-Aware (ALT):} Extends the minimal prompt with a
one-shot chain-of-thought exemplar \citep{Wei2022} to elicit explicit
reasoning over alternative utterances and their contextual appropriateness,
operationalizing the principle of Reasoning over Alternatives
(Section~\ref{sec:background}). The addition to the minimal prompt is
provided in Appendix~\ref{appxA}.
\item \emph{Knowledge-and-Motives-Aware (KMA):} Extends the minimal prompt with
an instruction to consider multiple plausible speaker knowledge states and
communicative motives before rating, operationalizing the principle of
Speaker Knowledge and Motives (Section~\ref{sec:background}). The addition
to the minimal prompt is provided in Appendix~\ref{appxA}.
\item \emph{Combined (COM):} Integrates both extensions above to test whether
jointly activating both pragmatic reasoning components yields improved
alignment with human ratings.
\end{itemize}

\section{Evaluation Metrics}
\label{sec:metrics}

Let $H$ and $M$ denote the human and model mean ratings, respectively,
for a given attribute, context, and utterance form. We assess alignment
at three levels.

\paragraph{Global pattern similarity.}
For each model--prompting condition pair, we measure overall alignment
across all $H$--$M$ pairs using three complementary metrics. The
\emph{Spearman rank correlation} ($\rho$) captures whether the model
preserves the relative ordering of human ratings across conditions,
without assuming a linear relationship. The \emph{Concordance
Correlation Coefficient} \citep[CCC;][]{Lin1989} 
jointly assesses co-variation and mean-level agreement.
The \emph{Root Mean Square Error} (RMSE) quantifies the average absolute
deviation between $H$ and $M$ on the original 7-point scale, providing
an interpretable measure of magnitude discrepancy.

\paragraph{Structural alignment.}
We assess whether models reproduce the direction of the ten significant
effects established in the human experiment (Section~\ref{sec:human}).
The \emph{Directional Agreement Score} (DAS) checks, for each of the
five significant main effects of form, whether the sign of the mean
difference $\Delta = M_{\text{precise}} - M_{\text{approximate}}$
matches the human direction: $\text{sign}(\Delta_M) =
\text{sign}(\Delta_H)$. The \emph{Interaction Sensitivity Score}
(ISS) applies the analogous check to the five significant form $\times$
context interactions, asking for each attribute whether the difference
in $\Delta$ between HP and LP conditions has the correct sign:
$\text{sign}(\Delta_M^{\text{HP}} - \Delta_M^{\text{LP}}) =
\text{sign}(\Delta_H^{\text{HP}} - \Delta_H^{\text{LP}})$.
Both scores range from 0 to 1, where 1 indicates perfect directional
agreement with the human benchmark across all relevant effects.

\paragraph{Magnitude calibration.}
Beyond directional agreement, we assess whether models reproduce the
\emph{magnitude} of the ten significant human effects. The \emph{Effect
Size Ratio} (ESR) is computed for each significant main effect and
form $\times$ context interaction separately:
\begin{equation}
  \text{ESR} = \frac{|\Delta_M|}{|\Delta_H|}
\end{equation}
where $\Delta = \bar{x}_{\text{precise}} - \bar{x}_{\text{approx}}$
for main effects, and $\Delta = (\Delta^{\text{HP}} -
\Delta^{\text{LP}})$ for interactions, with $\Delta^c =
\bar{x}_{\text{precise}}^c - \bar{x}_{\text{approx}}^c$ for context
$c$. ESR $= 1$ indicates perfect magnitude match; ESR $> 1$ indicates
exaggeration; ESR $< 1$ indicates attenuation. To summarize across
all ten effects, the \emph{Calibration Deviation Score} (CDS) is:
\begin{equation}
  \text{CDS} = \frac{1}{n}\sum_{i=1}^{n}|\,\text{ESR}_i - 1\,|
\end{equation}
where $i$ indexes the $n$ significant human effects (main effects and
interactions) with non-zero $|\Delta_H|$. Lower CDS indicates closer
alignment to human effect magnitudes overall.

By separating structural metrics (DAS, ISS) from magnitude metrics
(ESR, CDS), this framework enables principled assessment of both
\emph{which} inferences LLMs make and \emph{how strongly} they make
them.

\section{Results}
\label{sec:results}

\paragraph{Universal Structure, Variable Calibration.}
Structural alignment is uniformly high across all models and conditions:
DAS and ISS equal 1.0 for all attributes with non-zero human effects,
indicating perfect reproduction of both main effect polarity and
form $\times$ context interaction directions. Spearman $\rho$ values
range from 0.829 to 0.946, confirming strong rank-order correspondence
between model and human ratings across conditions.

\begin{figure*}[t!]
\centering
\includegraphics[width=\textwidth]{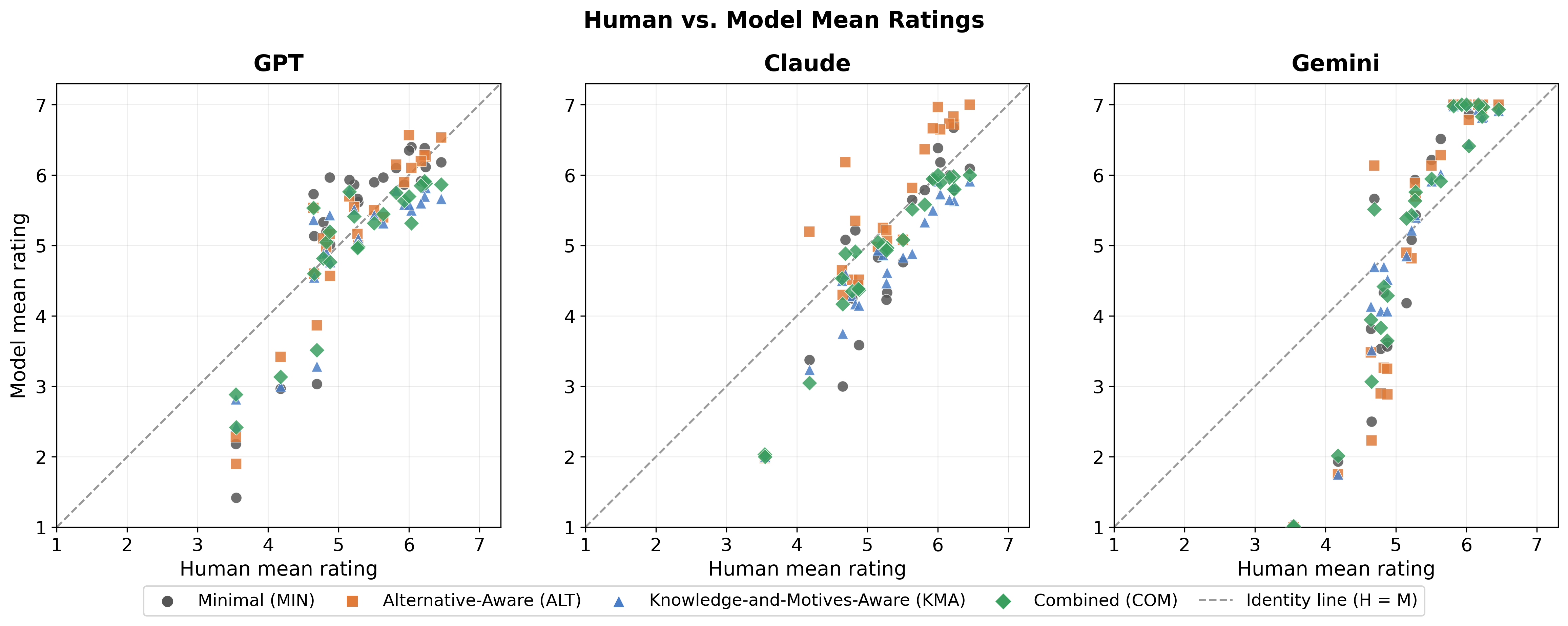}
\caption{Human vs.\ model mean ratings across all conditions (scenarios,
contexts, utterance forms, and social attributes) for each model and
prompting condition. Each point represents one human--model mean rating
pair; the dashed identity line ($H = M$) indicates perfect calibration.
Points above the line indicate model overestimation; points below indicate
underestimation. GPT clusters closely around the identity line across all
conditions, reflecting near-calibrated magnitude alignment. Claude shows
greater spread, with sensitivity to prompting condition visible in the
vertical displacement of individual condition clusters. Gemini displays
a characteristic compression along the x-axis with strong vertical
spread, reflecting the severe magnitude inflation reported in
Table~\ref{tab:cds}. Prompting conditions: MIN = Minimal (gray circles);
ALT = Alternative-Aware (orange squares); KMA = Knowledge-and-Motives-Aware
(blue triangles); COM = Combined (green diamonds).}
\label{fig:scatter}
\end{figure*}

However, the CCC and RMSE values in Table~\ref{tab:comprehensive_metrics}
\begin{table}[t!]
\centering
\small
\begin{tabular}{l l c c c}
\toprule
\textbf{Model} & \textbf{Prompt} & \textbf{Spearman $\rho$} & \textbf{CCC} & \textbf{RMSE} \\
\midrule
\multirow{4}{*}{GPT}
                & MIN & 0.856 & 0.703 & 0.811 \\
                & ALT & \textbf{0.880} & \textbf{0.812} & 0.565 \\
                & KMA & 0.829 & 0.754 & 0.601 \\
                & COM & 0.832 & 0.786 & \textbf{0.547} \\
\midrule
\multirow{4}{*}{Claude}
                & MIN & 0.849 & 0.730 & 0.766 \\
                & ALT & 0.847 & 0.754 & 0.724 \\
                & KMA & 0.920 & 0.739 & 0.701 \\
                & COM & \textbf{0.946} & \textbf{0.804} & \textbf{0.576} \\
\midrule
\multirow{4}{*}{Gemini}
                & MIN & 0.910 & 0.622 & 1.262 \\
                & ALT & 0.908 & 0.576 & 1.422 \\
                & KMA & 0.923 & \textbf{0.680} & \textbf{1.066} \\
                & COM & \textbf{0.932} & 0.659 & 1.125 \\
\bottomrule
\end{tabular}
\caption{Global pattern similarity metrics per model and prompting
condition. Spearman $\rho$ measures rank-order correspondence between
model and human mean ratings across all conditions; CCC
(Concordance Correlation Coefficient) jointly assesses co-variation
and mean-level agreement; RMSE reports average deviation on the
7-point scale. Prompting conditions: MIN = Minimal; ALT =
Alternative-Aware; KMA = Knowledge-and-Motives-Aware; COM = Combined.
Bold indicates the best value per model per metric.}
\label{tab:comprehensive_metrics}
\end{table}
reveal systematic calibration failures beneath this structural agreement.
CCC penalizes not only unsystematic noise but also systematic deviations
from the identity line $H = M$; the consistently lower CCC values
relative to Spearman $\rho$ therefore directly operationalize the
structure--magnitude dissociation: models preserve the \emph{ordering}
of human ratings while distorting their \emph{scale}. This is further
reflected in the RMSE values, which quantify the average deviation from
human ratings on the 7-point scale. Gemini shows the most severe
miscalibration (RMSE: 1.07--1.42), followed by Claude (RMSE: 0.58--0.77)
and GPT (RMSE: 0.55--0.81). Figure~\ref{fig:scatter} provides a global visualization of this structure--magnitude dissociation: while all models track the relative ordering of human ratings, systematic vertical displacement from the identity line reveals the degree of magnitude miscalibration for each model and prompting condition.

\paragraph{Magnitude Calibration Across Models.}
Table~\ref{tab:cds}
\begin{table}[t!]
\centering
\small
\begin{tabular}{l l c c c}
\toprule
\textbf{Model} & \textbf{Prompt} & $\textbf{CDS}_\textbf{m}$ & 
$\textbf{CDS}_\textbf{i}$ & \textbf{CDS} \\
\midrule
\multirow{4}{*}{GPT}
 & MIN & 0.325 & 0.460 & 0.393 \\
 & ALT & 0.303 & 0.392 & 0.348 \\
 & KMA & 0.360 & 0.444 & 0.402 \\
 & COM & \textbf{0.259} & \textbf{0.349} & \textbf{0.304} \\
\midrule
\multirow{4}{*}{Claude}
 & MIN & 0.845 & 0.231 & 0.538 \\
 & ALT & 1.235 & 0.205 & 0.720 \\
 & KMA & 0.454 & \textbf{0.167} & \textbf{0.310} \\
 & COM & \textbf{0.395} & 0.228 & 0.312 \\
\midrule
\multirow{4}{*}{Gemini}
 & MIN & 1.389 & 1.572 & 1.480 \\
 & ALT & 1.717 & 2.470 & 2.094 \\
 & KMA & \textbf{1.026} & \textbf{0.671} & \textbf{0.848} \\
 & COM & 1.241 & 1.434 & 1.338 \\
\bottomrule
\end{tabular}
\caption{Calibration Deviation Scores for main effects
($\text{CDS}_\text{m}$), interaction effects ($\text{CDS}_\text{i}$),
and their aggregate ($\text{CDS}$). Lower values indicate better magnitude
alignment with the human benchmark. Prompting conditions: MIN =
Minimal; ALT = Alternative-Aware; KMA = Knowledge-and-Motives-Aware; COM =
Combined. Bold indicates the best value per model per metric.}
\label{tab:cds}
\end{table}
reveals systematic architecture-dependent differences in magnitude
alignment, with a consistent dissociation between main-effect and
interaction calibration. 
\begin{figure*}[h!]
\centering
\includegraphics[width=\textwidth]{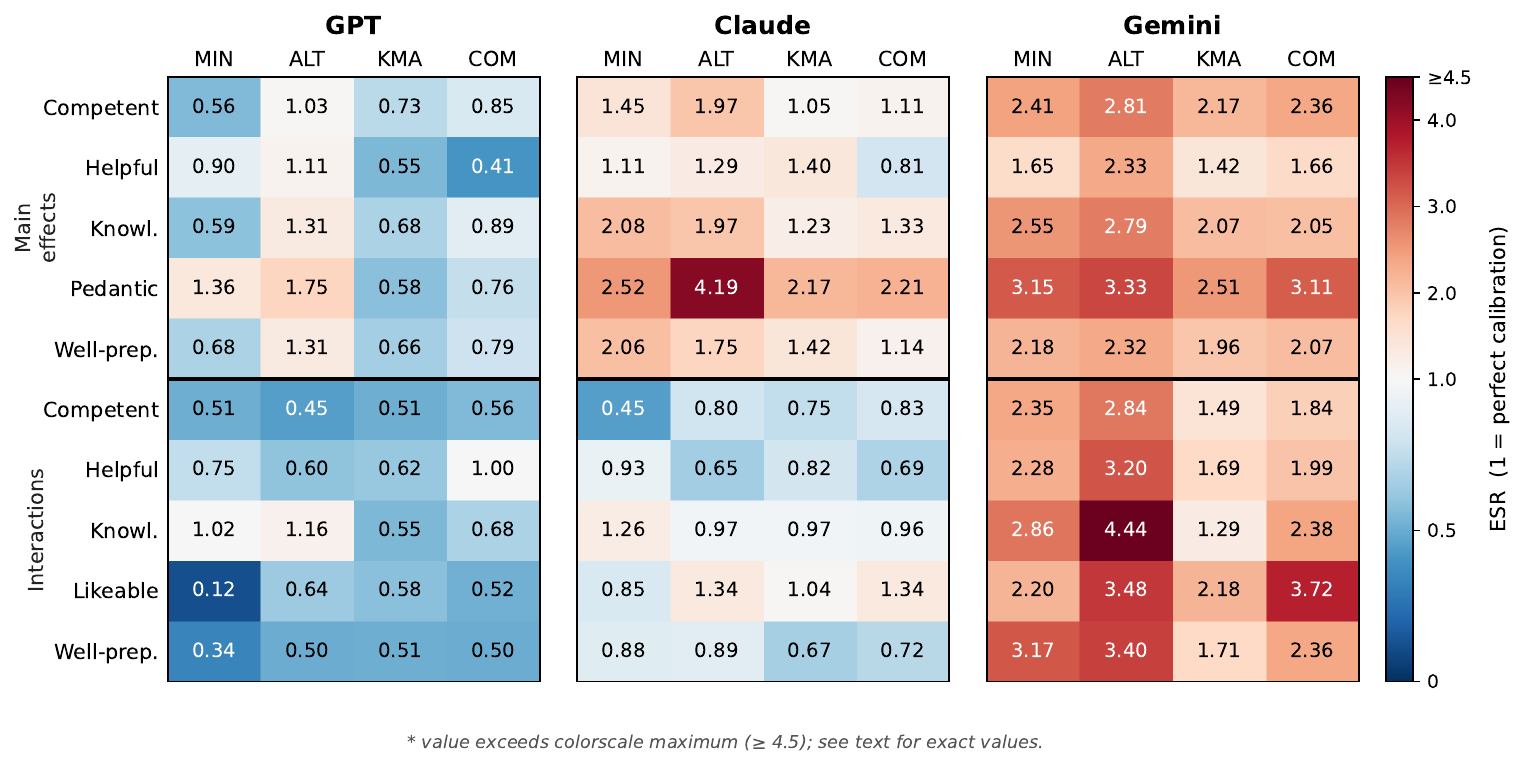}
\caption{Effect Size Ratios (ESR) per model, prompting condition, and
benchmark effect. Rows are grouped into main effects (top) and
form $\times$ context interactions (bottom); columns correspond to
prompting conditions (MIN, ALT, KMA, COM). Color encodes deviation
from perfect calibration (ESR $= 1$, white): blue indicates
attenuation, red indicates exaggeration. Values exceeding the
colorscale maximum of 4.5 are marked with an asterisk.}
\label{fig:esr_heatmap}
\end{figure*}
GPT shows the best overall calibration
(CDS: 0.3--0.4), with relatively modest deviations on both main
effects and interactions across all prompting conditions. Claude
exhibits moderate but uneven miscalibration: main-effect CDS varies
substantially across conditions (0.4--1.24), suggesting high
sensitivity to prompting, while interaction calibration is more stable
(0.17--0.23) and consistently lower than main-effect deviation ---
a pattern not shared by the other models. Gemini displays severe
magnitude inflation throughout, with interaction effects particularly
affected ($\text{CDS}_\text{i}$: 0.67--2.47), frequently exceeding
$2\text{--}3\times$ the human effect magnitude, while main-effect
miscalibration, though substantial ($\text{CDS}_\text{m}$:
1.03--1.72), is comparatively less extreme.

\paragraph{Prompting Effects on Calibration.}
The KMA condition produces the most consistent calibration improvements
across models prone to magnitude exaggeration. For Claude, CDS
decreases from 0.538 (MIN) to 0.310 (KMA), while for Gemini, CDS
decreases from 1.480 (MIN) to 0.848 (KMA), the largest absolute
reduction observed across any model--condition pair. GPT, already
well-calibrated at baseline, shows moderate sensitivity to prompting:
COM achieves the best overall CDS from 0.393 (MIN) to 0.304 (COM). 

Alternative-awareness prompting (ALT) produces mixed and sometimes
adverse effects. While it reduces main-effect deviation for GPT
($\text{CDS}_\text{m}$: 0.325 $\to$ 0.303) and 
interaction calibration for Claude ($\text{CDS}_\text{i}$: 0.231 $\to$ 0.205),
it substantially amplifies magnitude inflation for Gemini
($\text{CDS}_\text{i}$: 1.572 $\to$ 2.470), suggesting that
explicitly foregrounding alternative utterances may exacerbate
exaggeration in already poorly calibrated models.

Combined prompting (COM) stands out as the only condition that improves
all calibration-sensitive metrics (CCC, RMSE, $\text{CDS}_\text{m}$, $\text{CDS}_\text{i}$) relative to minimal prompting 
for every model.
For GPT, COM achieves
the best overall CDS (0.304) and lowest RMSE (0.547). For Claude, COM
produces the strongest global alignment across all three metrics
(Spearman $\rho$: 0.946, CCC: 0.804, RMSE: 0.576) and the best
$\text{CDS}_\text{m}$ (0.395), though KMA yields better interaction calibration
($\text{CDS}_\text{i}$: 0.167 vs.\ 0.228 under COM). For Gemini, COM
slightly reduces miscalibration relative to MIN (CDS: 1.480
$\to$ 1.338) while remaining less effective than KMA on CDS. The
consistent cross-model improvement under COM, even for GPT, which
shows little sensitivity to individual prompting components,
suggests that jointly activating both pragmatic reasoning processes
produces reliable alignment gains, even when the individual components
yield mixed results. However, the substantial gap between COM and KMA for Gemini's
calibration  indicates that the two components are not
fully additive, and that the chain-of-thought exemplar may partially
interfere with the epistemic uncertainty instructions for fine-grained
context sensitivity.

Figure~\ref{fig:esr_heatmap} provides a detailed view of these patterns
across all models, prompting conditions, and benchmark effects. For
Claude and Gemini, main effects are systematically exaggerated
(ESR $> 1$), while GPT shows near-calibrated or attenuated main
effects throughout. Interaction effects show more model-specific
variation across all three models. Gemini displays the strongest and
most consistent exaggeration overall, with several cells exceeding the
colorscale maximum. A left-to-right reduction in deviation is visible
for Claude and Gemini, reflecting the positive effect of the KMA
condition.

\section{Discussion}
\label{sec:discussion}

\paragraph{Structure Without Calibration.}
Our results demonstrate a systematic dissociation between structural and
quantitative alignment. All models achieve perfect directional agreement
(DAS = ISS = 1.0) and high rank correlations across all prompting
conditions, yet CCC values fall consistently below Spearman $\rho$,
and CDS reveals substantial magnitude deviations. This confirms that
LLMs reliably learn \emph{which} inferences arise from linguistic form
and context, while variably failing to reproduce \emph{how strongly}
those inferences operate. The pattern suggests that models acquire
directional pragmatic knowledge from training data, but do not
faithfully encode the graded, probabilistic character of human social
inference.

The architecture-dependent nature of calibration failure is noteworthy.
GPT approximates human effect magnitudes closely across all conditions,
while Gemini systematically inflates both main effects and interactions
--- sometimes by factors of 2--3. Claude occupies an intermediate
position but is highly sensitive to prompting, suggesting that its
default inference behavior is less stable. 
What drives these between-architecture differences remains an open 
question: since all three models are closed-source, their training 
objectives, fine-tuning procedures, and response normalization strategies 
are not publicly available, and we refrain from drawing strong causal 
conclusions from behavioral differences alone.

The cross-model consistency of COM has a further implication beyond 
model evaluation. If explicitly prompting for joint reasoning over 
alternatives \emph{and} speaker knowledge states is the only intervention 
that reliably improves calibration across all architectures, this 
suggests that human-like pragmatic inference may itself require both 
components to operate simultaneously, consistent with the RSA view 
that listeners engage in joint inference over utterance alternatives and 
latent speaker states \citep{FrankGoodman2012, GoodmanFrank2016}. 
Conversely, the adverse effects of ALT in isolation suggest that 
alternative-awareness without epistemic grounding may amplify rather
than moderate social inferences. This dissociation could be
investigated directly in human participants through paradigms that
selectively manipulate access to alternative utterances and speaker
context information.

\paragraph{Pragmatic Prompting and Its Limits.}
The prompting manipulations reveal a partial and asymmetric benefit of
pragmatically informed instructions. Explicitly prompting for speaker
knowledge states and motives (KMA) consistently reduces magnitude
deviation in overestimating models suggesting that directing attention
to epistemic uncertainty moderates the exaggeration of social
inferences. This is consistent with the theoretical view that pragmatic
meaning arises from reasoning over latent speaker states
\citep{GoodmanFrank2016, Bergen2016}, and that models benefit from
having this reasoning made explicit.

Alternative-awareness prompting (ALT), by contrast, produces mixed and
sometimes adverse effects. For GPT, it yields modest improvements on
both main-effect and interaction calibration. For Claude, it reduces
interaction deviation but simultaneously inflates main-effect deviation
to its highest value across all conditions, resulting in a net worse
overall calibration than minimal prompting. For Gemini, ALT produces
the worst calibration observed across any model--condition combination,
severely amplifying magnitude inflation relative to baseline. This
pattern suggests that explicitly foregrounding utterance alternatives,
without anchoring the reasoning in uncertainty about speaker states,
amplifies contrast effects rather than moderating them, most severely
in models with stronger baseline calibration deficits.

A notable finding is that combined prompting (COM) is the only
condition that improves all calibration-sensitive metrics relative to minimal
prompting across all three models simultaneously. This cross-model
consistency suggests that jointly activating reasoning over alternatives
and over speaker knowledge states produces a more robust pragmatic
inference process than either component alone. At the same time, COM clearly underperforms KMA on magnitude calibration for both models prone to exaggeration: for Claude, KMA
yields better interaction calibration, and for Gemini the advantage
of KMA over COM is even more pronounced, affecting both main-effect
and interaction calibration. This consistent pattern suggests that
the chain-of-thought exemplar introduced by ALT partially interferes
with the epistemic uncertainty instructions when both are combined,
and that this interference is more severe in models with stronger
baseline calibration deficits. The overall picture is one of reliable
directional improvement under COM, with remaining architecture-specific
trade-offs at the level of individual calibration components.

\paragraph{Limitations and Future Directions.}
The evaluation is grounded in a single experimental paradigm involving
numerical expressions across six scenarios and six social attributes.
Generalization to other pragmatic domains, such as scalar implicature,
politeness, or register variation, remains to be established. 
The human benchmark consists of condition means from a published 
behavioral study \citep{SoltEtAl2025}; by-participant variance is 
accounted for in the original study's statistical analysis, and our 
evaluation framework follows standard practice in LLM-as-participant 
work in operating at the level of condition means 
\citep{Argyle2023, Santurkar2023}.
We evaluate three proprietary frontier models; conclusions should not be
generalized to open-weight architectures or smaller models, which may
differ substantially in their pragmatic calibration. 

Prompting manipulations approximate the relevant pragmatic reasoning
mechanisms without constituting direct implementations. The prompts
activate reasoning processes that are theoretically motivated but not
formally equivalent to RSA inference; future work could examine whether
more explicit computational instantiations of alternative-based or
epistemic reasoning yield stronger calibration gains.

Comparing LLM condition means against individual human ratings rather 
than condition means yields consistently higher RMSE across all models 
and conditions (by 0.6--0.9 points on the 7-point scale), confirming 
that mean-level benchmarking is the more conservative measure and that 
reported calibration deviations are not an artifact of aggregation 
(see Appendix~\ref{appxB}).
Future work could explore whether fine-tuning on calibrated human judgment data yields more
robust alignment \citep{Ouyang2022}. The present study focuses
on inference-time interventions; training-time calibration objectives
remain an important direction for closing the structure--magnitude gap
identified here.

\section{Conclusion}
We investigated whether frontier LLMs approximate human social meaning
not only qualitatively but also quantitatively, grounding evaluation in
experimentally measured human effect sizes. Across three models and four
prompting conditions, all models reliably reproduce the directional
structure of human social inference, a finding that is robust across
architectures and prompting manipulations, while diverging substantially
in magnitude calibration. Pragmatically informed prompting partially
reduces these deviations, but its effects are architecture-dependent and
not uniformly beneficial. The ESR and CDS metrics introduced here provide
principled tools for diagnosing the structure--magnitude dissociation, and
we argue that separating directional fidelity from magnitude alignment is
a necessary step toward evaluating genuinely human-like social reasoning
in LLMs.

\section*{Ethics Statement}
This study evaluates proprietary LLMs via API under standard access
conditions. The human behavioral data was collected in a previously
published study \citep{SoltEtAl2025} following standard ethical
procedures for online behavioral research. Our findings concern
model-level tendencies in social attribute inference; we caution
against using automated social judgments of this kind in consequential
decision-making contexts without careful human oversight.


\section*{Data Availability}
The human behavioral data is reported in \citet{SoltEtAl2025} and is
publicly available at \url{https://osf.io/m4rhn} (experiment1.csv). 
The LLM rating data generated in this study
is publicly available at
\url{https://github.com/muehlenbernd/llm-social-calibration}.

\section*{Acknowledgements}

This work was funded by the Deutsche Forschungsgemeinschaft (DFG, German Research Foundation) – SFB 1412, 416591334. 

\bibliographystyle{lrec2026-natbib}
\section*{Bibliographical References}\label{sec:reference}
\bibliography{references}




\appendix

\section{Supplementary Materials: Appendices, Software, and Data}

\subsection{Prompt Texts}
\label{appxA}

The following example shows the minimal prompt of the scenario `bicycle' for the high precision context, the approximate numerical expression, and the social attribute competent.

{\small
\vspace{.3cm}\noindent\begin{tabular}{|p{7.2cm}|}
\hline
\textbf{Minimal Prompt:} \\
\hline
**Task description:** \\
In this experiment, you'll read a brief description of a situation involving two people. 
One of these people asks a question, and the second person answers it.
Your task will be to answer some questions about the second person. \\
\vspace{-.1cm}
**Task situation:**\\
Jamie's new bicycle was stolen. 
Fortunately it was insured, so Jamie has called the insurance company. \\
Insurance agent: "How much did the bicycle cost? I'll start the paperwork right away." \\
Jamie: "The bicycle cost about \$500." \\
\vspace{-.1cm}
**Task:**\\
Based on what Jamie says, how confident does Jamie sound? \\
\vspace{-.1cm}
Use a 7-point Likert scale: \\
1 = not at all confident \\
7 = very confident \\
\vspace{-.1cm}
Answer with a single number between 1 and 7. Give only the number, no other text.\\
\hline
\end{tabular}\vspace{.2cm}
}

For the Alternative-Aware condition, the minimal prompt is extended by a
one-shot chain-of-thought exemplar, inserted between the
\emph{Task Description} and \emph{Task Situation} blocks.

{\small
\vspace{.3cm}\noindent\begin{tabular}{|p{7.2cm}|}
\hline
\textbf{Alternative-Aware Prompt Extension:} \\
\hline
**Example situation:** \\
Jordan and Sam are planning a work meeting.\\
Jordan: "Do you know what time the meeting starts?"\\
Sam: "It starts at around 9."\\
\vspace{-.1cm}
**Example task:**\\
Based on what Sam says, how polite does Sam sound?\\
Use a 7-point Likert scale:\\
1 = not at all polite\\
7 = very polite\\
\vspace{-.1cm}
**Example reasoning (for illustration only):**\\
Sam gives an approximate answer rather than an exact time. In this context, an approximate answer can be appropriate and polite, since it provides useful information without unnecessary detail. Nothing in Sam's response suggests rudeness or disrespect.\\
\vspace{-.1cm}
**Example answer:** 6\\
\vspace{-.1cm}
**Now the actual task**\\
You will now see a new situation.\\
Please answer the question based only on the information given.\\
\hline

\end{tabular}\vspace{.2cm}
}

For the Knowledge-and-Motives-Aware condition, the \emph{Task} block
of the minimal prompt is replaced by an extended version that includes
explicit instructions to consider speaker knowledge states and
communicative motives prior to rating.

{\small
\vspace{.3cm}\noindent\begin{tabular}{|p{7.2cm}|}
\hline
\textbf{Knowledge-and-Motives-Aware Prompt:} \\
\hline
**Task:**\\
Based on what Jamie says, how competent does Jamie sound?\\
\vspace{-.1cm}
Before you answer, please note:\\
The same utterance can arise from different speaker knowledge states and motivations.
You should therefore avoid assuming a single motive or level of knowledge unless the context clearly supports it.\\
\vspace{-.1cm}
- Step 1: Briefly list two or three plausible reasons why the speaker might have chosen this wording,
considering both their possible knowledge state and their communicative goals.\\
- Step 2: Based on this uncertainty, provide a balanced social evaluation of the speaker.\\
\vspace{-.1cm}
Now it's your turn: How competent does Jamie sound?\\
Use a 7-point Likert scale:\\
1 = not at all competent\\
7 = very competent\\
\vspace{-.1cm}
Answer with a single number between 1 and 7. Give only the number, no other text.\\
\hline

\end{tabular}\vspace{.1cm}
}

The Combined condition integrates both extensions into the minimal prompt.
Since they target different blocks of the prompt structure, the two
additions can be inserted independently.

\subsection{Mean- vs.\ Individual-Level RMSE}
\label{appxB}

Table~\ref{tab:rmse_comparison} compares RMSE computed against human
condition means vs individual human ratings. 
Individual-level RMSE is substantially higher across all
models and conditions (by 0.5--0.9 points), confirming that
mean-level benchmarking is the more conservative measure and that
reported calibration deviations are not an artifact of aggregation.

\begin{table}[h!]
\renewcommand{\arraystretch}{0.81}
\centering
\small
\begin{tabular}{l l c c}
\toprule
\textbf{Model} & \textbf{Prompt} & \textbf{RMSE\textsubscript{mean}} & 
\textbf{RMSE\textsubscript{indiv}} \\
\midrule
\multirow{4}{*}{GPT}
 & MIN & 0.811 & 1.542 \\
 & ALT & 0.565 & 1.429 \\
 & KMA & 0.601 & 1.441 \\
 & COM & 0.547 & 1.420 \\
\midrule
\multirow{4}{*}{Claude}
 & MIN & 0.766 & 1.521 \\
 & ALT & 0.724 & 1.497 \\
 & KMA & 0.701 & 1.489 \\
 & COM & 0.576 & 1.433 \\
\midrule
\multirow{4}{*}{Gemini}
 & MIN & 1.262 & 1.822 \\
 & ALT & 1.422 & 1.937 \\
 & KMA & 1.066 & 1.691 \\
 & COM & 1.125 & 1.729 \\
\bottomrule
\end{tabular}
\caption{RMSE computed against human condition means
(\textbf{RMSE\textsubscript{mean}}) vs.\ individual human ratings
(\textbf{RMSE\textsubscript{indiv}}) per model and prompting condition.}
\label{tab:rmse_comparison}
\end{table}

\end{document}